\documentclass[11pt,twocolumn]{scrartcl}

\usepackage{casa_conf}	
\usepackage{graphicx}	

\usepackage{graphicx}
\usepackage{tabularx}
\usepackage{bbding}

\usepackage{xcolor}
\usepackage{soul}
\usepackage{outlines}
\usepackage{enumitem}

\usepackage{amsmath}

\definecolor{mypurple}{rgb}{0.85,0.7,1}
\definecolor{mygray}{rgb}{0.85,0.85,0.85}
\definecolor{mygreen}{rgb}{1,0.65,0.65}
\definecolor{mysand}{rgb}{0.9,0.8,0.5}

\newcommand{\quotes}[1]{``#1''}

\soulregister{\quotes}{1}
\soulregister{\ref}{7}
\soulregister{\cite}{7}
\soulregister{\citep}{7}
\soulregister{\citet}{7}
\soulregister{\subsection}{1}
\soulregister{\section}{1}
\soulregister{\label}{1}
\soulregister{\futurelet}{0}
\soulregister{\textbf}{1}

\usepackage{url}
\usepackage{array}
\usepackage[table]{xcolor} 
\usepackage{makecell}
\usepackage{csquotes}
\usepackage{textcomp}

\usepackage{algorithm}
\usepackage{algpseudocode}

\usepackage{etoolbox} 
\AtBeginEnvironment{algorithmic}{\footnotesize}

\usepackage[acronym]{glossaries}
\usepackage{bibunits}
\usepackage{float}
\usepackage{booktabs}
\usepackage{makecell}
\usepackage{multirow}

\newacronym{vh}{VH}{Virtual Human}

\newacronym{vlm}{VLM}{Vision–Language Model}
\newacronym{llm}{LLM}{Large Language Model}

\newacronym{eca}{ECA}{Embodied Conversational Agent}
\newacronym{gca}{GCA}{General Capable Agent}
\newacronym{npc}{NPC}{Non-Player Character}
\newacronym{ai}{AI}{Artificial Intelligence}
\newacronym{cd}{C-D}{Capability-Difficulty}

\usepackage{cuted}   
\usepackage{capt-of} 
\usepackage{multirow}
\usepackage{placeins}   
\usepackage{dblfloatfix} 

\usepackage{listings}
\newcommand{\orcidID}[1]{\textsuperscript{[ORCID:#1]}}

\title{A.D.A.M.O. (Agent for language-Driven Actions with Multimodal Observations): A Visual-Symbolic Framework for Virtual Humans}

\author{
Alessandro Emmanuel Pecora\orcidID{0009-0007-6826-5514}\thanks{These authors contributed equally.} \and
Stefano Calzolari\orcidID{0009-0009-7038-2212}\footnotemark[1] \and
Francesco Strada\orcidID{0000-0001-9197-9100} \and
Andrea Bottino\orcidID{0000-0002-8894-5089}\\
DAUIN, Politecnico di Torino, Torino, Italy \\ {\texttt\small\{name\}.\{surname\}@polito.it}
}

\begin{document}
\maketitle


\begin{strip}
\vspace*{-18mm} 
\centering
\includegraphics[width=\textwidth]{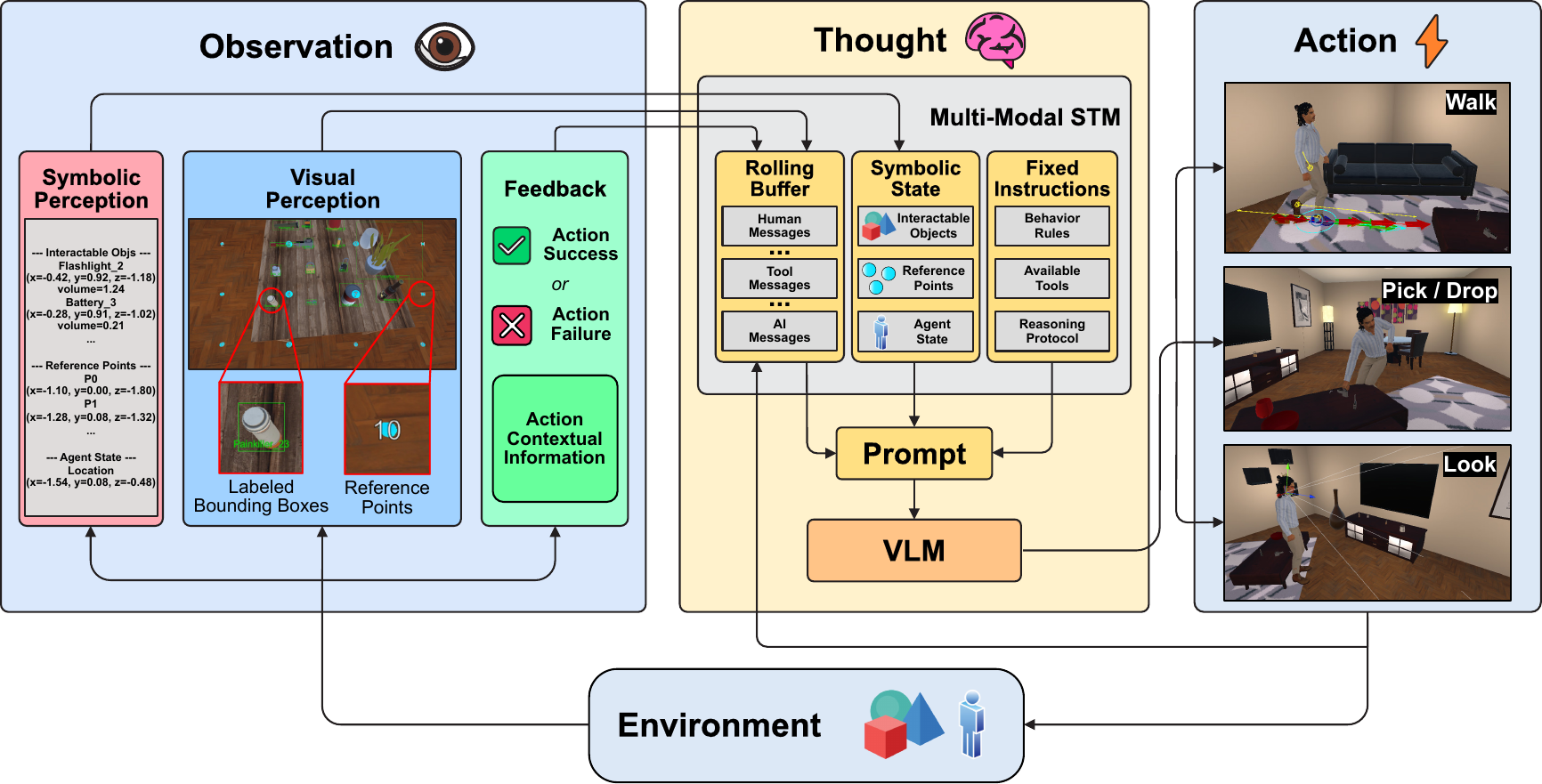}
\setlength{\emergencystretch}{3em}
\captionof{figure}{A.D.A.M.O. architecture follows a Perceive–Reason–Act loop. Observations combine symbolic state, egocentric visual input, and action feedback. A VLM with tool calling and multimodal short-term memory selects actions, which are executed as primitive operations (Walk, Pick/Drop, Look) and fed into the next cycle.}


\label{fig:teaser}
\vspace*{-4mm} 
\end{strip}


\begin{abstract}
Creating believable \glspl{vh} requires the coherent integration of perception, reasoning, and action mediated by language. A central challenge is to combine these components into a control loop grounded in interactive 3D environments. To this end, we present A.D.A.M.O. (Agent for language-Driven Actions with Multimodal Observations), a visual-symbolic framework for language-driven \glspl{vh} that leverages a pretrained \gls{vlm} with tool calling to unify perception, reasoning, and action within a single control loop. A.D.A.M.O. maintains a dual visual-symbolic world model that combines egocentric visual input and synchronized symbolic state to support grounded task-oriented behavior from natural language prompts. To support diagnostic evaluation, we introduce a controlled task suite organized by a \gls{cd} taxonomy that breaks down spatial tasks into procedural and linguistic complexity. Experiments in controlled scenes show that semantic labeling strongly influences task completion and failure modes, reducing perceptual ambiguity while shifting failures toward downstream execution, whereas reasoning errors remain comparatively rare.
\end{abstract}
\linebreak
\linebreak
\linebreak
\keywords{Virtual Humans, Agentic AI, Vision Language Model, Embodied AI, Spatial Reasoning, Cognitive Architecture}
\section{Introduction}
\label{sec:intro}

\glsresetall

Creating believable \glspl{vh} -- artificial characters that look and act like humans in simulated environments~\cite{lugrin2021introduction} -- remains a central challenge at the intersection of computer graphics and embodied AI. Beyond visual realism, believability requires agents to connect multimodal perception of 3D environments with goal-directed action through coherent reasoning. When such grounding is achieved, \glspl{vh} can exhibit human-like behavior in virtual worlds, enabling applications in training, interactive entertainment, and collaborative environments \cite{yang2025embodied,guo_developing_2023}. However, current  \glspl{vh} systems still rely on extensive manual scripting and domain-specific design, which limits flexibility and makes grounded behavior difficult to scale across tasks and environments.

Natural language provides a natural interface for specifying goals and interacting with \glspl{vh}, yet existing approaches often remain only partially integrated. \glspl{eca} support rich language-driven interaction but typically lack perceptual grounding in 3D environments, limiting situated interaction, such as spatial reference and object manipulation \cite{saga2025greta,kopp2022fabric}. Conversely, classical \glspl{gca} achieve robust control through reinforcement learning but often operate over symbolic states with limited linguistic and visual grounding \cite{team2021open}. Recent LLM-driven embodied agents integrate language and action via tool use and memory \cite{wang2024jarvis,wang2025karma}, but tightly coupling visual observations, symbolic reasoning, and embodied control in dynamic 3D settings remains challenging.

To address this gap, we introduce A.D.A.M.O. (Agent for Language-Driven Actions with Multimodal Observations), a visual-symbolic framework for language-driven \gls{vh} control that unifies perception, reasoning, and embodied action within a single decision loop. A.D.A.M.O. builds on tool-calling and agentic paradigms \cite{yao2022react,schick2023toolformer} and integrates two complementary representations: a visual stream, where egocentric renderings processed by a pretrained \gls{vlm} encode spatial and semantic context, and a synchronized symbolic stream that captures object properties and relations as structured text. Cross-referencing these streams supports grounded multimodal reasoning and tool-based interaction, providing a system-level alternative to fully handcrafted control pipelines in controlled 3D environments.

This paper contributes a \glspl{vh}-specific system-level integration of language-based decision making, multimodal perception, short-term memory, and tool-based action, together with an interpretable diagnostic evaluation framework for studying language-driven embodied behavior in \gls{vh}. To support this evaluation, we introduce a controlled task suite organized by a \gls{cd} taxonomy that decomposes spatial tasks by procedural and linguistic complexity. Using this setup, we analyze semantic scaffolding, task difficulty, and failure modes under controlled observability. Our experiments show that semantic labeling is a major factor in task completion and failure distribution, reducing perceptual ambiguity while shifting failures toward downstream execution effects, whereas reasoning-related errors remain comparatively rare. Code and data are available at our project page~\footnote{https://github.com/CGVGroup/ADAMO}.
\section{Related Work}
\label{sec:related_works}


\noindent\textbf{Virtual Agents.}
The landscape of virtual agents reveals a persistent challenge in integrating linguistic interaction with embodied behavior, evident across different generations of agent architectures.
Traditional Game AI for \glspl{npc} relies on rule-based behaviors \cite{uludaugli2023non,millington2019ai}, achieving reliable yet manually scripted interactions that require extensive authoring to generalize.
Research then diverged into two specialized directions.
\glspl{eca} prioritized linguistic sophistication, achieving human-like interaction through coordinated verbal and nonverbal behavior \cite{saga2025greta,knob_arthur_2024,calzolari2025toward}, yet decoupling communicative intent from physical realization. 
Recent \gls{llm}-enhanced systems \cite{saga2025greta,llanes2024developing} improve dialogue but remain disconnected from spatial reasoning and object manipulation.
Conversely, \glspl{gca} prioritized embodied behavior, leveraging \glspl{llm} to decompose goals and acquire skills in sandbox environments \cite{fan2022minedojo,wang2023voyager}. 
Most rely on symbolic world states offering limited perceptual grounding, while recent vision language variants \cite{wang2024jarvis,zhao2024steve,zhao2024see} still lack robust contextual grounding.
Even attempts to bridge these dimensions fall short: language-driven \glspl{npc} \cite{rao2024collaborative} remain heavily user-dependent, lacking autonomous perception-action loops for independent spatial reasoning.

Regarding spatial awareness, early systems such as Max~\cite{kopp2003max} introduced symbolic control for environment-aware interaction but relied on handcrafted rules, which limited adaptability.  Li et al.~\cite{li2025exploring} later proposed virtual humans operating from text-encoded scene descriptions, enabling rich linguistic reasoning over spatial relations but requiring exhaustive manual annotation.  These omniscient setups lack scalability to complex environments and cannot exploit visual cues for disambiguation~\cite{pecora2025survey}.

A.D.A.M.O. addresses these limitations by unifying egocentric vision and structured symbolic memory within a single language-based control layer.  
The symbolic stream encodes spatial structure for explicit reasoning, while the visual stream supports perceptual verification and disambiguation.  Leveraging pretrained \glspl{vlm}, the framework interprets high-level commands and autonomously composes primitive actions, reducing reliance on fully handcrafted pipelines and supporting grounded perception–reasoning–action in controlled virtual environments.

\noindent\textbf{Embodied AI Evaluation.}
Beyond virtual agents, robotics and embodied \gls{ai} pursue similar goals -- grounded perception, linguistic understanding, and goal-directed action -- in physical or simulated settings~\cite{liu2025aligning,wang2025karma,ramanathan2019nadine}. 
Although operating under different constraints (partial observability, noisy sensing, physical dynamics), the embodied AI community has developed unified benchmarks to evaluate language-conditioned task execution.  
Frameworks such as ALFRED~\cite{shridhar2020alfred} emphasize procedural execution, while more recent efforts like LoTa-Bench~\cite{choi2024lota} and OpenEQA~\cite{majumdar2024openeqa} incorporate richer language understanding but rely on binary success metrics that obscure task difficulty and error sources. 
As a result, it remains difficult to identify where agents fail, which reasoning skills are most demanding, or how to systematically scale task complexity.  
Comparable evaluation frameworks are still missing in \gls{vh} research, despite the need to jointly assess linguistic, spatial, and embodied competence.  
A.D.A.M.O. addresses this gap through a \gls{cd} taxonomy that decomposes tasks along two orthogonal axes -- procedural capability (\textit{what to do}) and linguistic complexity (\textit{how it is described}) -- offering interpretable, scalable measures of embodied reasoning difficulty that explain not only whether agents succeed, but also \emph{why} they struggle on specific task dimensions.

\section{Methods}
\label{sec:methods}

Our goal is to equip \glspl{vh} with \emph{language-driven control} that integrates perception, reasoning, and action within a coherent decision loop in controlled 3D environments. In this setup, the agent receives natural language instructions, reasons over spatial and semantic context, and selects goal-directed actions aligned with the task.

This section describes the design of A.D.A.M.O.\ through two complementary layers. The \textbf{logical architecture} (Sec.~\ref{sec:logical_architecture}) defines how the agent perceives, reasons, and selects actions from multimodal inputs under partial observability. The \textbf{infrastructure architecture} (Sec.~\ref{sec:infrastructure_architecture}) details the system-level implementation that supports this loop, separating high-level reasoning from real-time simulation and action execution.

\subsection{Logical Architecture}
\label{sec:logical_architecture}

In embodied \gls{ai}, agents must integrate perception, reasoning, and action to operate continuously within their environment. Following robotics literature \cite{murphy2019introduction,kopp2022fabric}, we adopt the \emph{perceive--reason--act} paradigm as our architectural foundation, instantiated through the ReAct framework \cite{yao2022react} and adapted here to a multimodal, language-driven \gls{vh} setting. In A.D.A.M.O., reasoning (\emph{Thought}), action (\emph{Act}), and perception (\emph{Observation}) form a closed feedback loop. The simulation engine maintains full access to the world state, while the agent operates under partial and sequential observability through egocentric perception exposing only task-relevant local context at each step (Fig.~\ref{fig:teaser}). 

\noindent\textbf{Thought.}  
Reasoning is implemented through a \gls{vlm} with tool-calling capabilities that, at each iteration, generates a structured \emph{reasoning note} followed by a specific tool selection. The reasoning note serves as a short-term sketchpad that summarizes perceptual evidence and selects the next action, functioning as an explicit intermediate reasoning trace in the spirit of chain-of-thought prompting \cite{wei2022chain} and making the decision process interpretable within the Perceive--Reason--Act loop.

\noindent\textbf{Act.}  
Actions are executed as tool invocations that translate linguistic intent into embodied control. Tools interact directly with the simulation engine, which enforces physical consistency (e.g., automatic repositioning when targets are out of reach), thereby decoupling high-level reasoning from low-level execution details. Tool execution produces new observations that update the world state and close the control loop. In our setup, the agent uses a minimal set of primitive tools -- \texttt{Look}, \texttt{Walk}, \texttt{Pick}, and \texttt{Drop} -- covering perception, navigation, and object manipulation. This core interface supports the experiments presented in this paper, and its modular design allows additional primitives to be incorporated in future extensions.

\noindent\textbf{Observation.}  
Observations integrate \textbf{perceptual observation} (\textit{what does the agent sense?}) and \textbf{feedback} (\textit{what happened during this action?}). Perceptual observation provides a multimodal view of the environment by combining egocentric visual input with a synchronized symbolic state. \textit{Visual perception} consists of an RGB frame augmented with object tags, bounding boxes for interactable entities, and sparse tagged reference points projected into the camera view, which serve as stable 3D spatial anchors. \textit{Symbolic perception} mirrors this view as structured text, encoding the currently available interactable object list (using the same tags as the visual stream), along with 3D coordinates, object volumes, reference-point positions, and the \emph{agent state}, which stores the current agent pose to preserve spatial consistency across reasoning iterations.

Execution feedback is returned as a concise textual summary of the last tool outcome (e.g., success, failure, or automatic correction), providing contextual signals that support coherent reasoning in subsequent iterations.

\noindent\textbf{Multimodal Short-Term Memory (MSTM).}  
To preserve temporal and multimodal coherence across the Thought--Act--Observation loop, A.D.A.M.O. maintains a Multimodal Short-Term Memory (MSTM) that unifies recent visual inputs, symbolic state, and interaction history within a single prompt. The MSTM follows a message-based ChatML structure \cite{openai2023chatml}, which organizes dialogue into explicit roles (\emph{system}, \emph{AI}, \emph{tool}, \emph{human}), and comprises (i) fixed instructions (\emph{system}) defining behavioral rules and tool semantics, (ii) a dynamically updated symbolic state storing objects, spatial references, and agent pose (\emph{system}), and (iii) a rolling buffer that retains the last $k$ \textit{human} (initial prompt and egocentric images), \textit{AI} (reasoning notes and tool calls), and \textit{tool} messages (returning environmental observations, i.e., perceptual updates and feedback). All prompts are provided in the project repository.

\subsection{Infrastructure Architecture}
\label{sec:infrastructure_architecture}
\begin{figure}[t]
    \centering
\includegraphics[width=0.75\columnwidth]{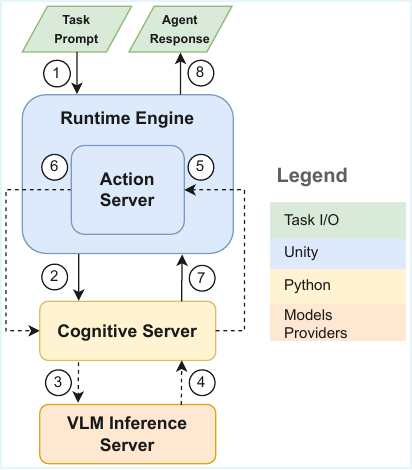}
    \caption{Infrastructure architecture. Numbered labels indicate dataflow and execution sequence across modules; dotted steps (3–6) implement the Thought–Act–Observation cycle.}
    \label{fig:infra-architecture}
\end{figure}

The infrastructure architecture (Fig.~\ref{fig:infra-architecture}) implements the Thought--Act--Observation loop through a modular system that decouples high-level reasoning from simulation and action execution. This separation supports flexibility and interoperability without entangling cognitive decisions with real-time environment dynamics.

The \textbf{Runtime Engine}, implemented in Unity, manages the 3D environment and agent embodiment, handling rendering, physics, and I/O. It captures perceptual data and executes embodied actions through an embedded \textbf{Action Server} that translates high-level tool calls into executable Unity operations and returns both feedback and updated observations.

The \textbf{Cognitive Server}, implemented in Python, orchestrates the Thought--Act--Observation loop, managing prompt construction, memory updates, and tool invocations. Reasoning and action selection are delegated to a \textbf{VLM Inference Server}, which provides access to pretrained VLMs or LLMs through a unified interface.

At runtime, a task prompt and initial observations are sent to the Cognitive Server, which iteratively queries the VLM for the next action, executes it through the Action Server, and updates the Multimodal Short Term Memory (MSTM)

with new observations until task completion (Fig.~\ref{fig:infra-architecture}).

\noindent\textbf{Execution Flow.}
As shown in Fig.~\ref{fig:infra-architecture}, the Runtime Engine receives a task prompt and initial observations (1) and forwards them to the Cognitive Server (2), which assembles a multimodal prompt and queries the VLM Inference Server (3). The VLM returns either a tool invocation or a termination decision (4). Tool calls are executed via the Action Server in the Unity Runtime (5), which updates the environment and returns perceptual observations and execution feedback (6). These updates are integrated into the MSTM, enabling iterative Thought--Act--Observation cycles (3–6) until completion, after which the Cognitive Server signals termination (7) and the Runtime Engine delivers the final agent response (8).

\section{Experiments}
\label{sec:experiments}

We evaluate A.D.A.M.O. on spatial reasoning and manipulation tasks in a controlled simulated 3D environment, aiming to analyze performance across task types and sensitivity to key design choices in perception and reasoning.

\begin{figure}[th!]
    \centering
    \includegraphics[width=\columnwidth]{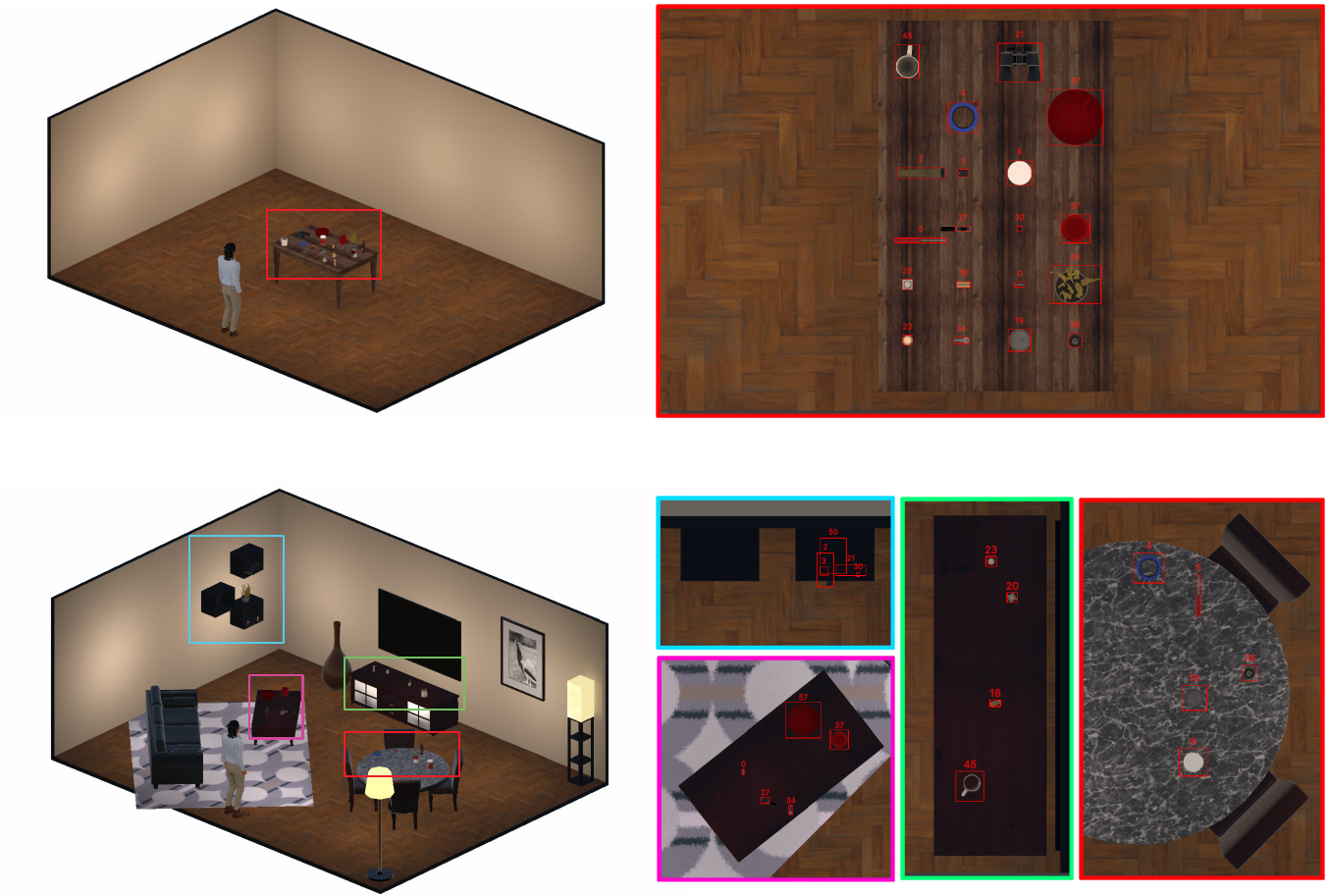}
    \caption{Scene in our benchmark (top left: \textbf{S1}; bottom left: \textbf{S2}) with closeups of the respective objects arrangement (right).}
    \label{fig:scenes}
\end{figure}

Our experimental setup includes 15 self-contained \emph{episodes}, each defined by a triplet $(S, T, C)$: the \textbf{scene} $S$ specifies the 3D environment, the interactable objects layout, and initial agent pose; the \textbf{task prompt} $T$ is a natural-language instruction describing an object-centric manipulation task, and the \textbf{checker} $C$ deterministically evaluates task completion. We consider two scenes (Fig.~\ref{fig:scenes}): a simple tabletop scene where all interactable objects are placed on a single table (S1) and a living-room scene with the same objects distributed across multiple pieces of furniture, yielding a more complex spatial layout (S2), both containing the same 19 interactable objects but differing in spatial arrangement and contextual complexity. Task prompts combine basic manipulation primitives (\textit{pick}, \textit{place}) with varying spatial constraints (\textit{near}, \textit{on}).

For each episode, the checker returns the completion rate, a normalized score in $[0,1]$ based on the fraction of satisfied atomic conditions (e.g., placing one of four required objects yields a score of $0.25$). Task difficulty is characterized using the agent-agnostic \gls{cd} taxonomy described in Sec.~\ref{subsec:C-D-taxonomy}. Full details on scenes, and checker logic are provided in the online repository. Tasks are listed in Table \ref{tab:task_prompts}.

To study sensitivity to architectural design choices, we compare variants of the perceptual and reasoning stack that differ in (i) the \gls{vlm} backbone and (ii) the object-labeling scheme used for visual and symbolic references. Specifically, we test \texttt{GPT-4o-vision} (G4O) and \texttt{Claude-Sonnet-3.5} (S35) under identical inputs, and contrast a \textbf{semantic} labeling scheme (\texttt{SEM}), where tags encode class and instance information, with an \textbf{opaque} one (\texttt{OPAQ}) using numeric identifiers only. This design allows us to examine how semantic scaffolding affects grounding performance and action efficiency. 

\glsreset{cd}
\subsection{\gls{cd} Taxonomy}
\label{subsec:C-D-taxonomy}

We use a \emph{Capability--Difficulty (C-D) taxonomy} as a simple, interpretable heuristic to organize task difficulty within our experimental setup. Each episode, defined by a natural-language instruction, is decomposed into three unit types: \textbf{Actions} \textcolor{blue}{ACT}, which count explicit manipulation primitives; \textbf{References} \textcolor{green}{REF}, which capture the effort required to uniquely identify target objects through linguistic or perceptual descriptors; and \textbf{Relations} \textcolor{red}{REL}, which denote spatial constraints that must be satisfied in the final configuration. Each unit is assigned an integer cost reflecting its relative complexity, and the C-D level of an episode is given by the sum of its unit costs.

This taxonomy provides an agent-agnostic ordering of task difficulty that supports diagnostic analysis of performance trends and failure modes across task families. Locomotion and perceptual actions are treated as implicit and are not explicitly scored. Table~\ref{tab:c-d-taxonomy} summarizes the units and costs used in our experiments; full task specifications and cost assignments are provided in the online repository.

\noindent\textbf{Example.}
In a scene with a \emph{yellow ball}, a \emph{black cup}, and a \emph{white fork}, the task \textcolor{blue}{Pick} the \textcolor{green}{black object} and \textcolor{blue}{place} it \textcolor{red}{near} the \textcolor{green}{spherical object} yields \textcolor{blue}{ACT}=2 (pick+drop), \textcolor{green}{REF}=3 (color+shape), and \textcolor{red}{REL}=2 (proximity), for a total C-D level of $7$.

\begin{table}[t!]
\centering
\scriptsize
\caption{Capability–Difficulty (C–D) taxonomy: units, specification, and levels.}
\label{tab:c-d-taxonomy}
\renewcommand{\arraystretch}{1.15}
\setlength{\tabcolsep}{6pt}
\begin{tabular}{p{0.05\linewidth} l p{0.45\linewidth} c}
\hline
Dim & Unit & Specification & Level \\
\hline
\noalign{\hrule height 0.9pt}
\multirow{2}{*}{\centering\textbf{Act}}
 & pick & Pick an object. & 1 \\
 & drop & Release or place an object. & 1 \\
\noalign{\hrule height 0.9pt}
\multirow{9}{*}{\centering\textbf{Ref}}
 & id            & Numerical identifier, unique in scene. & 0 \\
 & tag           & Semantic tag or name. & 1 \\
 & attrColor     & Color attribute. & 1 \\
 & attrShape     & Shape attribute. & 2 \\
 & groupSimple   & Single-criterion group filter. & 2 \\
 & attrPattern   & Pattern attribute. & 3 \\
 & superExtremum & Superlative over an attribute: min or max. & 3 \\
\noalign{\hrule height 0.9pt}
\multirow{3}{*}{\centering\textbf{Rel}}
 & floor & Place on the floor. & 1 \\
 & prox  & Place near another object. & 2 \\
 & area  & Place within a designated area or region. & 4 \\
\noalign{\hrule height 0.9pt}
\end{tabular}
\end{table}

\subsection{Instrumentation}
\label{sec:metrics}

For each \gls{vlm} and labeling scheme, every episode is executed 15 times (with a 10-minute cap) to account for model stochasticity and runtime failures. Results are reported as episode-level averages and aggregated by scene, \gls{vlm}, and labeling scheme. We measure completion rate (CR), model usage (number of tool calls and per-call generation time), and control activity (counts of \emph{Walk}, \emph{Look}, \emph{Pick}, and \emph{Drop} actions). All agentic models are evaluated under identical settings. The MSTM rolling buffer stores the last $k{=}50$ interaction steps. Decoding uses greedy sampling (temperature $=0.0$, top-$p{=}0.0$) with a maximum output length of $4000$ tokens per message. Visual observations use a $1920{\times}1080$ RGB frame and a $6{\times}6$ grid of raycasted reference points uniformly distributed across the agent’s field of view. Tasks are ordered for increasing C-D levels, which range between 1 for T1 until 20 for T15 (Table \ref{tab:task_prompts}).

\begin{table*}[h!]
\footnotesize
\definecolor{my_gray}{RGB}{220,220,220}
\rowcolors{2}{white}{my_gray}
\centering
\caption{Full list of task prompts and corresponding \gls{cd} levels of our taxonomy.}
\label{tab:task_prompts}
\begin{tabularx}{\textwidth}{c X c c c c}
\toprule
\textbf{Task ID} & \textbf{Task Prompt} & \textbf{ACT} & \textbf{REF} & \textbf{REL} & \textbf{TOT} \\
\midrule
T1  & Pick the object tagged as 34                                                                                                        & 1  & 0 & 0 & 1  \\
T2  & Pick the plate                                                                                                                      & 1  & 1 & 0 & 2  \\
T3  & Pick the blue circular object                                                                                                       & 1  & 1 & 0 & 2  \\
T4  & Pick the plate and place it on the floor                                                                                            & 2  & 1 & 1 & 4  \\
T5  & Pick the object tagged as 4 and place it near the indoor plant                                                                      & 2  & 1 & 2 & 5  \\
T6  & Pick the smallest object then place it on the floor                                                                                 & 2  & 3 & 1 & 6  \\
T7  & Pick the flashlight and place it on the floor in a corner of the room                                                               & 2  & 1 & 4 & 7  \\
T8  & Pick the glass and the plate then put them on the floor                                                                             & 4  & 2 & 2 & 8  \\
T9  & Pick object tagged as 37 then place it next to the plate, then go pick object tagged as 30 and place it next to object tagged as 3 & 4  & 1 & 4 & 9  \\
T10 & Pick the glass then place it next to the plate, moreover pick the binoculars and place it next to the screwdriver                   & 4  & 4 & 4 & 11 \\
T11 & Pick object tagged as 4 and place it near the indoor plant, then pick the glass and place it next to the plate                      & 4  & 3 & 4 & 12 \\
T12 & Pick the cup, peanut butter, and pills and place them on the floor                                                                  & 6  & 3 & 3 & 12 \\
T13 & Pick the cylindrical can and place it next to the indoor plant, and pick the binoculars and place it next to the screwdriver        & 4  & 4 & 4 & 12 \\
T14 & Pick all objects with white parts and place them on the floor                                                                       & 10 & 3 & 5 & 15 \\
T15 & Pick all objects that have a printed product label and place them on the floor                                                      & 10 & 5 & 5 & 20 \\
\bottomrule
\end{tabularx}
\end{table*}

\section{Results}
\label{sec:results}

We analyze A.D.A.M.O.’s behavior on spatial reasoning and manipulation tasks to examine performance under controlled variation, identify dominant failure modes, and assess sensitivity to key design choices in perception and reasoning.

\subsection{Model and Labeling Scheme}

Table~\ref{tab:performance} reports average performance (completion rate, walk, look and total tool calls, and call execution time) across all tasks and scenes, aggregating 15 repetitions per episode for each backbone (G4O, S35) and labeling scheme (SEM, OPAQ). Across both models, the labeling scheme is the dominant factor: semantic labeling (SEM) markedly increases completion relative to opaque labeling (OPAQ), while backbone choice primarily affects efficiency and the way failures manifest.

\textbf{Backbone effects.}
Averaging results across labeling schema, G4O achieves higher average completion than S35 ($0.68$ vs.\ $0.61$) and lower per-call latency ($6.85$s vs.\ $8.01$s), but executes more actions per episode ($4.70$ vs.\ $4.19$). This difference reflects distinct failure behaviors: S35 more frequently terminates early, inferring task completion before all goals are satisfied, whereas G4O tends to maintain the reasoning--action loop until environment feedback confirms completion. Thus, the backbone appears to modulate decisiveness versus completion persistence more than overall task resolvability.

\textbf{Effect of semantic labeling.}
SEM labeling yields the largest gains, increasing completion from $0.48{\to}0.88$ (+83\%) for G4O and from $0.38{\to}0.83$ (+118\%) for S35, while reducing exploratory perception. The number of \textit{Look} actions drops by $12\%$ for G4O and $41\%$ for S35, and total action counts decrease across both models, suggesting more direct, plan-driven trajectories under semantic grounding.

\textbf{Failure modes and exploration.}
A quantitative error breakdown clarifies these trends. Specifically, we parsed execution logs from failed episodes to categorize failure causes at scale. Under OPAQ, failures are overwhelmingly perceptual (83.5\% against 14.2\% of execution-related), reflecting ambiguity in object identification without semantic cues. Under SEM, perceptual errors drop to 65.8\%, while execution-related failures increase to 31.1\%, suggesting a shift from misidentification toward downstream action and termination effects. Reasoning-related errors remain rare in both settings (2--3\%).

\begin{table}[t!]
\definecolor{my_gray}{RGB}{220, 220, 220}
\rowcolors{2}{white}{my_gray}
\centering
\footnotesize
\caption{Average performances across all episodes for labeling scheme (top blocks) and, for SEM scheme only, per-scene (bottom blocks). 
}
\label{tab:performance}
\resizebox{\columnwidth}{!}{%
\begin{tabular}{*{6}{c}}
\toprule
\textbf{Model} &
\textbf{CR ↑} &
\textbf{\makecell{Time per\\Call (s) ↓}} &
\textbf{\#Walk} &
\textbf{\#Look} &
\textbf{\#Total} \\
\midrule

\hiderowcolors
\multicolumn{6}{c}{\textbf{Labeling Scheme: OPAQ}} \\[-2pt]
\cmidrule(lr){1-6}
\showrowcolors
G4O & \textbf{0.48} & \textbf{7.63} & 0.53 & 1.61 & 4.80 \\ 
S35 & 0.38 & 8.31 & 0.38 & 2.10 & 4.57 \\ 
\midrule
\hiderowcolors
\multicolumn{6}{c}{\textbf{Labeling Scheme: SEM}} \\[-2pt]
\cmidrule(lr){1-6}
\showrowcolors
G4O & \textbf{0.88} & \textbf{6.06} & 0.62 & 1.41 & 4.60 \\ 
S35 & 0.83 & 7.72 & 0.23 & 1.25 & 3.80 \\ 
\midrule

\hiderowcolors
\multicolumn{6}{c}{\textbf{Scene: S1 (SEM Labeling Scheme)}} \\[-2pt]
\cmidrule(lr){1-6}
\showrowcolors
G4O & \textbf{0.93} & \textbf{4.71} & 0.00 & 1.08 & 3.67 \\ 
S35          & 0.90          & 7.94 & 0.19 & 1.19 & 3.75 \\ 
\midrule
\hiderowcolors
\multicolumn{6}{c}{\textbf{Scene: S2 (SEM Labeling Scheme)}} \\[-2pt]
\cmidrule(lr){1-6}
\showrowcolors
G4O & \textbf{0.83} & \textbf{7.12} & 1.24 & 1.74 & 5.53 \\ 
S35          & 0.76          & 7.49 & 0.28 & 1.30 & 5.89 \\ 
\bottomrule
\end{tabular}
}%
\end{table}

\begin{figure*}[h!]
    \centering
    \includegraphics[width=1.00\textwidth]{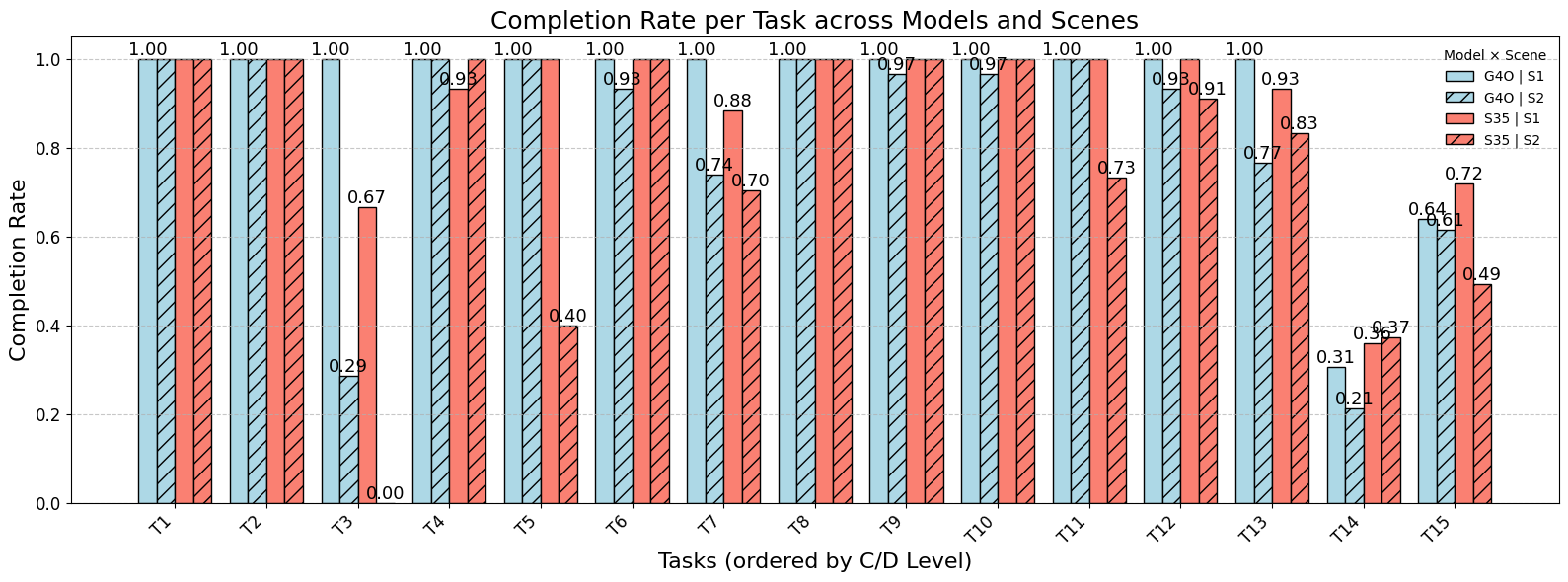}
    \caption{Completion rate across all tasks for different scenes and models using SEM labeling scheme.}

    \label{fig:all_tasks}
\end{figure*}

\subsection{Tasks and Scenarios}
\label{subsec:task_and_scenarios}

Given the strong performance difference between the two labeling conditions, we next analyze how scene complexity and task structure affect performance under semantic labeling. Figure~\ref{fig:all_tasks} reports per-task completion across both models and environments, while Table~\ref{tab:performance} summarizes scene-level averages.

\textbf{Scene effects.}
Although S1 (tabletop) and S2 (living room) contain the same 19 objects, S2 introduces greater spatial dispersion and visual clutter. Completion rates are consistently higher in S1 (G4O: $0.93$, S35: $0.90$) than in S2 (G4O: $0.83$, S35: $0.76$). The more complex layout in S2 increases exploratory behavior: \textit{Walk} actions rise from near zero in S1 to $1.24$ (G4O) and $0.28$ (S35) in S2, with corresponding increases in \textit{Look} actions. G4O’s broader exploration partially compensates for reduced visibility, yielding higher completion at the cost of additional actions.

\textbf{Task difficulty and task-family trends.}
Tasks naturally cluster into families with increasing procedural and perceptual demands. \emph{Single-object manipulation} tasks (T1–T5, C-D $\leq 5$) achieve near-perfect completion, indicating reliable performance on basic grounding tasks. A notable exception is T3 (\textit{``pick the blue circular object''}), where color–shape discrimination stresses perceptual grounding and reveals differences in how models balance visual evidence and pretrained semantic priors.


\emph{Comparative and relational} tasks (e.g., T6: \textit{``smallest object''},
T10: \textit{``next to the screwdriver''}) and \emph{moderate multi-object} tasks (T8--T13, C-D $\in [6,12]$) remain highly reliable, despite requiring attributes and global spatial relations that are not explicitly encoded in the semantic labels (e.g., T7: \textit{``flashlight in a corner''}, where the target location is not associated with any semantically labeled object). These results suggest that, even without explicit semantic labels supporting comparative and relational task structure, such tasks can still be handled through the complementarity of the dual stream, with visual and symbolic representations jointly providing the information needed to resolve object comparisons and relations.

Performance degrades for \emph{attribute-heavy} and \emph{long-horizon} task families (T14–T15, \mbox{C-D $\geq 14$).} T14 (\textit{``all objects with white parts''}) exposes conflicts between pretrained category–attribute priors and visual evidence, while T15 (\textit{``printed product labels''}) combines fine-grained attribute recognition with extended action sequences, resulting in the lowest completion rates. These effects are amplified in more cluttered scenes.

Across all tasks under SEM labeling, and considering all trials and all models, we observe a significant negative rank correlation between C-D levels and completion rate (Spearman $\rho = -0.53$, $ p < 0.05$), indicating that increasing task complexity tends to place greater demands on multimodal grounding.

\subsection{Ablation Studies}
\label{sec-appendix:ablation}

This section examines the sensitivity of key architectural and inference hyperparameters on A.D.A.M.O.’s performance by varying one parameter at a time while keeping all others fixed. All ablations are performed on Scene S1 using the G4O backbone with the SEM labeling scheme, and report completion rate across all 15 tasks along with selected efficiency metrics.

\textbf{MSTM Rolling Buffer Size ($k$).}
Varying the MSTM buffer size controls the amount of recent multimodal context available to the agent. As shown in Table~\ref{tab:ablation_mstm_k}, completion improves from $0.82$ at $k{=}4$ to $0.86$ at $k{=}10$, and peaks at $0.93$ for the baseline $k{=}50$. Smaller buffers increase exploratory behavior, with more \texttt{Look} actions and model calls (e.g., $6.34$ calls at $k{=}4$ vs.\ $4.67$ at $k{=}50$), indicating compensation for truncated context through repeated perception. Larger buffers preserve temporal coherence at the cost of higher latency.

\begin{table}[h]
\definecolor{my_gray}{RGB}{220,220,220}
\rowcolors{2}{white}{my_gray}
\centering
\footnotesize
\renewcommand{\arraystretch}{1.15}

\caption{MSTM rolling-buffer size.}
\label{tab:ablation_mstm_k}

\begin{tabular}{lcccc}
\toprule
\textbf{k} 
& \textbf{CR ↑}
& \textbf{\# Look}
& \makecell{\textbf{\# Model}\\\textbf{Call ↓}}
& \makecell{\textbf{Time per}\\\textbf{Call (s) ↓}} \\
\midrule
4             & 0.82 & 1.27 & 6.34 & 3.94 \\
6             & 0.84 & 1.20 & 9.20 & \textbf{3.85} \\
10            & 0.86 & 1.47 & 6.99 & 4.05 \\
50\textsuperscript{\textdagger} & \textbf{0.93} & \textbf{1.08} & \textbf{4.67} & 4.71 \\
\bottomrule
\end{tabular}
\end{table}

\begingroup
\renewcommand{\thefootnote}{}
\footnotetext{\textdagger\ Baseline configuration used for comparison throughout all the ablation studies.}
\endgroup

\textbf{VLM Sampling Hyperparameters.}
Table~\ref{tab:ablation_sampling} shows that deterministic decoding is most reliable for structured tool use: the quasi-deterministic setting achieves the highest completion rate ($0.93$). Moderate stochasticity leads to a sharp drop ($0.82$), while high stochasticity partially recovers performance ($0.85$), suggesting that increased output diversity occasionally helps, but overall degrades tool-execution reliability.

\begin{table}[h]
\definecolor{my_gray}{RGB}{220, 220, 220}
\rowcolors{2}{white}{my_gray}
\centering
\footnotesize
\caption{
VLM sampling configurations.
}
\label{tab:ablation_sampling}
\resizebox{\columnwidth}{!}{%
\begin{tabular}{lcc}
\toprule
\textbf{Configuration} & \textbf{(Temp ; Top-p)} & \textbf{CR ↑} \\
\midrule
Quasi-deterministic\textsuperscript{\textdagger}
& (0.0 ; 0.0) 
& \textbf{0.93} \\
Moderately stochastic 
& (0.3 ; 0.5) 
& 0.82 \\
Highly stochastic 
& (0.8 ; 1.0) 
& 0.85 \\
\bottomrule
\end{tabular}
}%
\end{table}

\textbf{3D Reference-Point Density.}
Spatial anchoring improves performance up to a point. Increasing reference-point density from $3{\times}3$ to $6{\times}6$ raises completion to $0.93$ while reducing model calls (4.67) and tokens (32,977) (Table~\ref{tab:ablation_refs}). Further densification to $7{\times}7$ lowers completion to $0.87$, consistent with increased visual and symbolic clutter, which may dilute salient geometric cues and make action selection less reliable.

\begin{table}[h]
\definecolor{my_gray}{RGB}{220, 220, 220}
\rowcolors{2}{white}{my_gray}
\centering
\footnotesize
\caption{
Number of 3D reference points. 
}
\label{tab:ablation_refs}
\resizebox{\columnwidth}{!}{%
\begin{tabular}{l c c c} 
\toprule
\textbf{\# Ref. Points} 
& \textbf{CR ↑}
& \textbf{\# Model Call ↓}
& \makecell{\textbf{Tokens} \\ \textbf{per Ep. ↓}}  \\
\midrule
3x3 & 0.92 & 5.71 & 42963.89 \\ 
4x4 & 0.90 & 5.48 & 41023.82 \\
5x5 & 0.89 & 5.77 & 44816.31 \\ 
6x6\textsuperscript{\textdagger} & \textbf{0.93} & \textbf{4.67} & \textbf{32977.61} \\ 
7x7 & 0.87 & 5.34 & 43372.84 \\ 
\bottomrule
\end{tabular}
}
\end{table}

\textbf{Image Resolution.}
Higher input resolution improves perceptual grounding but increases latency (Table~\ref{tab:ablation_resolution}). Completion rises from $0.76$ at $640{\times}360$ to $0.93$ at $1920{\times}1080$, while per-call time increases from $2.65$ s to $4.71$ s, highlighting a clear performance--efficiency trade-off in the perception stage.

\begin{table}[h]
\definecolor{my_gray}{RGB}{220, 220, 220}
\rowcolors{2}{white}{my_gray}
\centering
\footnotesize
\caption{
Input image resolution. 
}
\label{tab:ablation_resolution}
\resizebox{\columnwidth}{!}{
    \begin{tabular}{l c c}
    \toprule
    \textbf{Resolution} & \textbf{CR ↑} & \makecell{\textbf{Time per} \\ \textbf{Call (s) ↓}}\\
    \midrule
    $640\times 360$ & 0.76 & \textbf{2.65} \\
    $1280 \times 720$ & 0.85 & 3.52 \\
    $1920 \times 1080$\textsuperscript{\textdagger} & \textbf{0.93} & 4.71 \\
    \bottomrule
    \end{tabular}
}
\end{table}

\subsection{Limitations and Future Work}

A.D.A.M.O. is currently evaluated in controlled simulated environments, where the agent operates under partial and sequential observability through egocentric perception. While this setup supports systematic analysis and still includes occlusions and action-induced scene changes, it does not explicitly model perceptual noise, exogenous environment dynamics, or uncertainty-aware action selection. The framework also relies only on multimodal short-term memory, without persistent spatial or episodic memory. This limits coherence in extended interactions and makes long-horizon tasks more challenging. From a modeling perspective, the current implementation depends on VLMs with native tool-calling support, which restricts compatibility with part of the open-source ecosystem. In addition, fine-grained visual attribute grounding remains difficult, as pretrained semantic priors can conflict with visual evidence in attribute-sensitive tasks. Most importantly, semantic labeling is a major factor in task completion and failure distribution. Although it substantially improves performance, it still requires lightweight scene-side authoring and highlights the current dependence of the framework on semantic scaffolding. The present results should therefore not be interpreted as evidence that raw visual grounding alone is sufficient for reliable language-driven virtual human control. Rather, they point to a trade-off between performance and manual semantic scaffolding, showing that strong performance can already be achieved with lightweight scene-side authoring. Future work will extend the framework with persistent memory, broaden support to open-source VLMs, further reduce manual semantic scaffolding through automatic semantic extraction, and evaluate the system in more complex settings, including dynamic environments, multi-agent scenarios, and human--VH collaboration.
\section{Conclusion}

We presented A.D.A.M.O., a visual-symbolic framework for language-driven virtual humans that integrates multimodal perception, reasoning, and embodied control within a unified decision loop. By combining egocentric visual input with structured symbolic state and tool-based actions, A.D.A.M.O. supports interpretable, language-mediated behavior in controlled 3D environments.

Using a Capability--Difficulty-based evaluation protocol, we showed that semantic labeling is a major factor in reliable performance, substantially reducing perceptual ambiguity and shifting failures toward downstream execution effects, while reasoning-related errors remain comparatively rare. Task difficulty and fine-grained visual attributes emerge as the main bottlenecks under increased complexity. Experimental results show that A.D.A.M.O. offers a system-level framework and diagnostic setup for studying grounded language-driven behavior in virtual humans, while helping identify the current bottlenecks across perception, memory, and action execution.


\bibliographystyle{unsrt}
\bibliography{main}

\end{document}